\documentclass{article}
\usepackage{spconf,amsmath,graphicx,booktabs,amssymb,multirow,subfigure}
\usepackage{enumitem}
\usepackage{hyperref}
\usepackage{caption}
\usepackage{pifont}

\usepackage{xcolor}

\title{ENHANCING CROSS-DOMAIN DETECTION: ADAPTIVE CLASS-AWARE CONTRASTIVE TRANSFORMER}
\name{Ziru Zeng, Yue Ding, Hongtao Lu$^{\ast}$ \thanks{$\ast$ Corresponding author: Hongtao Lu, also with MOE Key Lab of Artificial Intelligence, AI Institute, Shanghai Jiao Tong University, Shanghai, China, htlu@sjtu.edu.cn. This paper is supported by NSFC (No.62176155), Shanghai Municipal Science and Technology Major Project (2021SHZDZX0102).}}
\address{Department of computer science and engineering, Shanghai Jiao Tong University, China}
\begin{document}
\ninept
\maketitle
\begin{abstract}
Recently, the detection transformer has gained substantial attention for its inherent minimal post-processing requirement. However, this paradigm relies on abundant training data, yet in the context of the cross-domain adaptation, insufficient labels in the target domain exacerbate issues of class imbalance and model performance degradation. To address these challenges, we propose a novel class-aware cross domain detection transformer based on the adversarial learning and mean-teacher framework. First, considering the inconsistencies between the classification and regression tasks, we introduce an IoU-aware prediction branch and exploit the consistency of classification and location scores to filter and reweight pseudo labels. Second, we devise a dynamic category threshold refinement to adaptively manage model confidence. Third, to alleviate the class imbalance, an instance-level class-aware contrastive learning module is presented to encourage the generation of discriminative features for each class, particularly benefiting minority classes. Experimental results across diverse domain-adaptive scenarios validate our method’s effectiveness in improving performance and alleviating class imbalance issues, which outperforms the state-of-the-art transformer based methods.
\end{abstract}

\section{Introduction}
\label{sec:intro}
In recent years, the object detection model based on transformer architecture has become a very potential detector with its complete end-to-end characteristics, no post-processing and better generalization performance \cite{tvt,daformer}. However, when testing in a scenario with domain gaps, its performance will still be significantly degraded due to different data distributions and domain shifts, such as weather changes and stylistic variations. This motivates us to explore the unsupervised domain adaptation based on transformer detector.
\begin{figure}[ht]
    \centering
    \setlength{\abovecaptionskip}{-5pt}
    \includegraphics[scale=0.25]{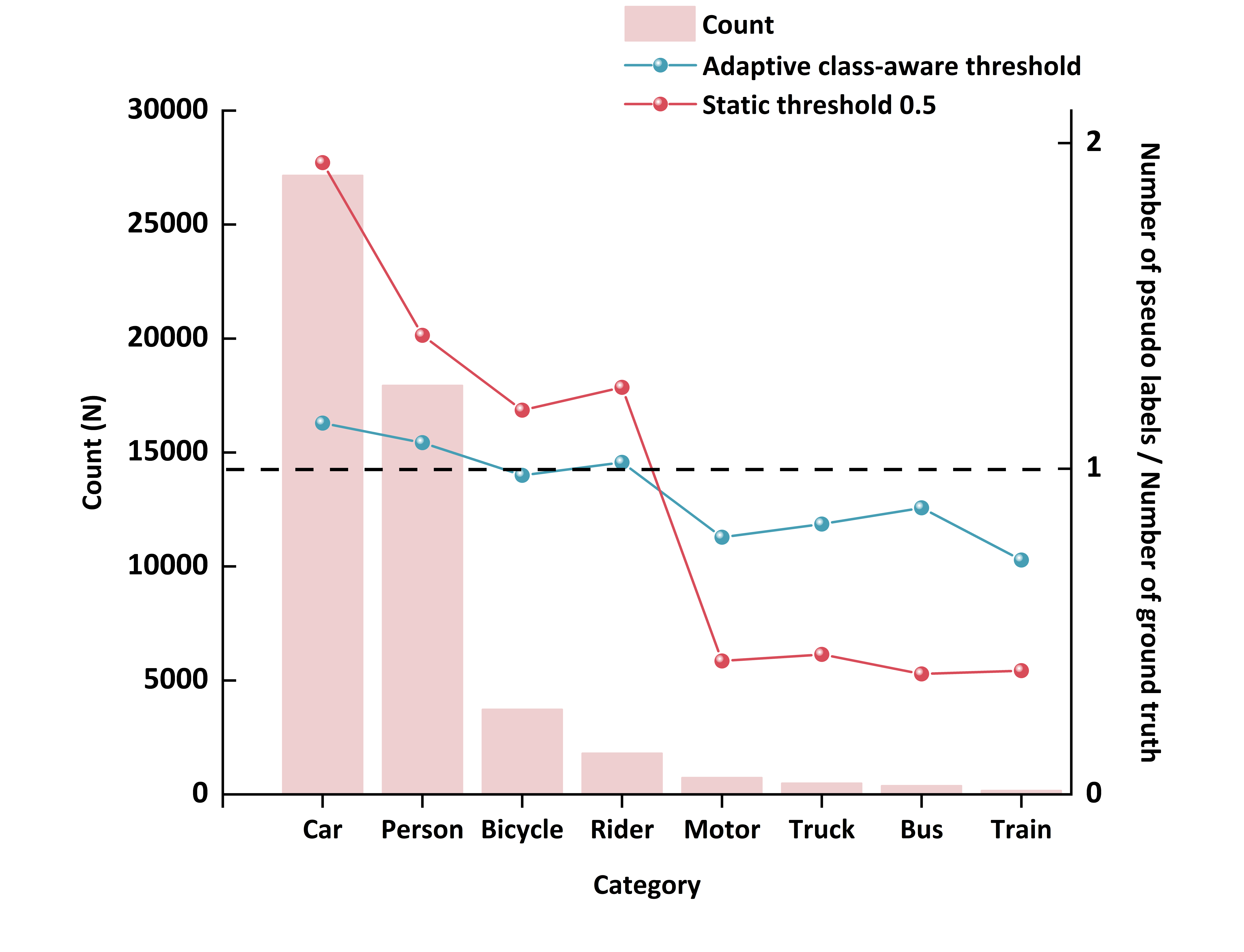}
    \caption{We quantify the count of the distinct objects in the foggy cityscapes training set and display the ratio between pseudo labels and ground truth under our adaptive class-aware threshold and unified static threshold respectively.}
    \label{fig:imbalance}
    \vspace{-0.85cm}
\end{figure}

\begin{figure*}[htbp]
    \centering
    \setlength{\abovecaptionskip}{-2pt}
    \includegraphics[scale=0.2]{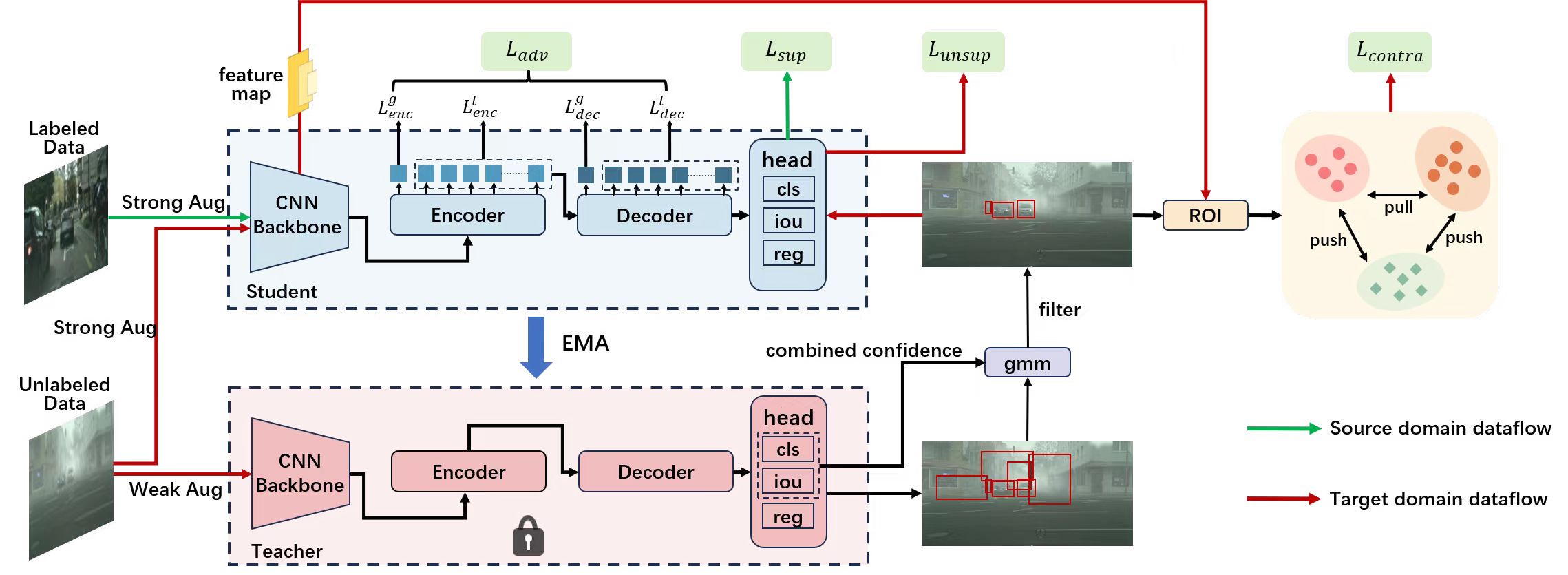}
    \caption{Overview of the proposed framework. In burn-up stage, we freeze the teacher model's dataflow and conduct adversarial feature learning to train student model. In mutual-learning stage, the teacher model is activated to generate pseudo labels for weak-aug target domain images. We use GMM model with combined confidence to obtain the class threshold and filter the boxes. Then we exploit the ROIAlign to extract the feature of filtered boxes and compute the contrastive loss via contrastive learning module.}
    \label{fig:overall model}
    \vspace{-0.7cm}
\end{figure*}
Unsupervised domain adaptation aims to align a labeled source domain’s detector with an unlabeled target domain, bridging the domain gap \cite{SWDA,cross}.In cross-domain detection, we have the whole training images and complete labels of the source domain, while only have the image of the target domain. Almost all existing transformer-based detectors for domain adaptation apply adversarial learning to extract common features between different domains \cite{SFA,AQT}. Nevertheless, owing to the absence of instance labels for target domain, method based adversarial learning fails to exploit the latent category information in the target domain, still manifesting an obvious performance gaps compared to the fully supervised models. Inspired by the success of recent study \cite{MTTrans,cross-modal,enhanced} using self-training method, we link mean-teacher mechanism \cite{mean} with adversarial learning to generate pseudo labels for target domain, thus mining the information inside the data of target domain. Despite this effort, there still exists some obstacles in mean-teacher framework hindering the enhancement of detection performance.

Existing work \cite{MTTrans} based on vanilla mean-teacher architecture filters the pseudo labels only using their classification scores with a unified static threshold. However, it ignores the inconsistency \cite{consistent} between classification and regression task in object detection, which means high confidence score do not always guarantee the localization accuracy \cite{soft}. Simply using classification scores to filter pseudo labels may leave the inaccurate pseudo boxes, which inevitably introduces bias into the detection model and leads to performance degradation. Besides, the prior study overlooked the bias issue introduced by class imbalance. Owing to the imbalance data distribution, the model tends to predict the majority classes while assigning lower confidence scores to minority classes.  As shown in Fig \ref{fig:imbalance}, taking foggy cityscapes \cite{foggy} dataset as an example, there is a serious category imbalance issue in data distribution. Using uniform threshold across all categories will increase the false positives as the number of pseudo labels in majority classes, like car and person, far surpasses the count of ground truth. As a result, it will lead to inefficient training and biased supervision, amplifying the class imbalance issue.

To overcome these obstacles, we propose the adaptive class-aware contrastive transformer model. To deal with inaccurate localization for pseudo boxes, we introduce an IoU prediction branch to refine and reweight pseudo-labels by leveraging both classification and localization scores. Furthermore, to address the issue of class imbalance, a Gaussian Mixture Model (GMM) \cite{semidetr, consistent} is implemented to generate adaptive threshold for each category, utilizing the scores of each category pseudo boxes. This module not only enable to stabilize the number of true positive samples for all classes pseudo labels at different training steps but also serve the parameters for subsequent category reweighting. Moreover, to further mitigate class imbalance issue, we introduce a class-aware contrastive learning module to enhance the similarity of the same class instances in terms of the feature spatial distribution and ensures the considerable spatial separation between different class instances, particularly for minority classes by reweighting.

The main contributions of this paper are as follows: (1) We apply the consistency of classification and localization to filter and reweight pseudo labels, thereby acquiring higher quality pseudo boxes for robust training. (2) We introduce the GMM module derived adaptive category thresholds, which can reduce the false negatives for minority classes to mitigate biased supervision due to data class imbalance. (3) Design of an object-level class-aware contrastive learning module, which enhances feature separability specially for rare categories by reweighting and avoid overfitting of majority classes. (4) The proposed method shows exceptional performance across three representative domain-adaptive benchmarks, surpassing prior transformer-based methods.

\section{METHOD}
\label{sec:format}
As shown in Fig \ref{fig:overall model}, motivated by \cite{MTTrans}, we combine feature adversarial learning and self-training methods to form a multi-stage model training framework. In the first stage, we only train the student model using domain adversarial learning to aligns domain distributions. In the second stage, we initialize first-stage model weights into teacher-student model. The teacher model generates pseudo labels for student training, while the student model transfers the knowledge to the teacher model with Exponential Moving Average (EMA) \cite{mean} method. Detailed loss functions for these two stages are outlined in Section \ref{ssec:loss}.
\subsection{Preliminaries}
\label{ssec:subhead}
\textbf{Query-Token Wise Alignment.} To maximize self-training's potential while addressing low-quality pseudo-labels due to domain shifts, we introduce query-token alignment inspired by SFA \cite{SFA} . We employ domain adversarial training by adding gradient reversal layers and domain discriminators after both global and local tokens, which extracts domain-invariant features. The loss for each domain discriminator is expressed by the following formula: 
\vspace{-0.2cm}
\begin{equation}
    L(f) = -\frac{1}{N}\sum_{i=1}^{N}[d\log{D(f^{i})}+(1-d)\log{(1-D(f^{i}))}],
    \vspace{-0.2cm}
\end{equation}
where $d$ denotes the domain label (1 for source domain, 0 for target domain). $D$ represents the domain classifier, while $f$ signifies the token feature fed into the classifier. $N$ is set to 1 when inputting the global token, otherwise it equals the number of local tokens. The total adversarial loss is as follows: 
\vspace{-0.18cm}
\begin{equation}
    L_{adv}=\lambda_{g}(L(f_{enc}^{g})+L(f_{dec}^{g}))+\lambda_{l}(L(f_{enc}^{l})+L(f_{dec}^{l})),
    \vspace{-0.18cm}
\end{equation}
where $\lambda_{g}$ is coefficient for global tokens, and $\lambda_{l}$ is for local tokens.\\
\textbf{Mean-Teacher Framework.} Building upon the initial model through the above adversarial learning, we use teacher model to predict the weak augmented target images and apply EMA method \cite{mean} to update Teacher model. This process enables us to supervise the student network in target domain. The student model loss can be written as:
\vspace{-0.2cm}
\begin{equation}
    \begin{aligned}
        L_{s} &= L_{sup}+\lambda_{unsup} L_{unsup} \\
          &= L_{det}(X_{s},Y_{s}) + \lambda_{unsup}L_{det}(X_{t},\hat{Y_{t}}), \\
    \end{aligned}
    \vspace{-0.2cm}
\end{equation}
Where $X_{s}$ and $Y_{s}$ represent the source domain image and labels, respectively. $X_{t}$ is strong augmented target domain and $\hat{Y_{t}}$ are the filtered pseudo labels predicted by the teacher model. 

\begin{table*}[t] 
\centering
\footnotesize
\setlength{\tabcolsep}{8pt} 
\setlength{\abovecaptionskip}{0cm}
\caption{Results on weather adaptation, from cityscapes to foggy cityscapes, FRCNN and DefDETR are abbreviations for Faster R-CNN and Deformable DETR, respectively. "\dag" denotes that the method use VGG-16 backbone, otherwise use ResNet-50.}
\label{table1}
\begin{tabular}{l|l|llllllll|l}
\toprule
Methods & Detector &person &rider &car &truck &train &bus &mcycle &bicycle& mAP\\
\midrule
SWDA\cite{SWDA} & FRCNN &49.0 &49.0 &61.4 &23.9 &22.9 &43.1 &31.0 &45.2 &40.7 \\
ViSGA\cite{VisGA} & FRCNN &38.8 &45.9 &57.2 &29.9 &\textbf{51.9} &50.2 &31.9 &40.9 &43.3 \\
PT\dag\cite{PT} & FRCNN &40.2 &48.8 &59.7 &30.7 &30.6 &51.8 &35.4 &44.5 &42.7 \\
MGA\dag\cite{MGA} &FRCNN &45.7 &47.5 &60.6 &31.0 &44.5 &52.9 &29.0 &38.0 &43.6 \\
\midrule
EPM\cite{EPM} &FCOS &44.2 &46.6 &58.5 &24.8 &29.1 &45.2 &28.6 &34.6 &39.0 \\
KTNet\dag\cite{KTNet} &FCOS &46.4 &43.2 &60.6 &25.8 &40.4 &41.2 &30.7 &38.8 &40.9 \\
\midrule
SFA\cite{SFA} &DefDETR &46.5 &48.6 &62.6 &25.1 &29.4 &46.2 &28.3 &44.0 &41.3 \\
MTTrans\cite{MTTrans} &DefDETR &47.7 &49.9 &65.2 &25.8 &33.8 &45.9 &32.6 &46.5 &43.4 \\
DA-DETR\cite{DA-DETR} &DefDETR &49.9 &50.0 &63.1 &24.0 &37.5 &45.8 &31.6 &46.3 &43.5 \\
AQT\cite{AQT} &DefDETR &49.3 &52.3 &64.4 &27.7 &46.5 &\textbf{53.7} &36.0 &46.4 &47.1 \\
\midrule
ACCT(Ours) &DefDETR &\textbf{53.6} &\textbf{58.9} &\textbf{69.4} &\textbf{31.1} &33.7 &53.5 &\textbf{42.6} &\textbf{54.4} &\textbf{49.6} \\
\bottomrule
\end{tabular}
\label{table:table1}
\vspace{-0.6cm}
\end{table*}
\subsection{IoU-guided Pseudo Label Refinement}
\label{ssec:subhead}
Previous research \cite{harmonious}  highlights limitations when using only classification scores for pseudo box filtering, leading to suboptimal model performance. Most existing studies \cite{soft,refine} are based on two-stage training or post-processing such as Non-Maximum Suppression (NMS) \cite{faster} to measure box uncertainty. However, these above approaches are difficult to combine with the end-to-end and post-processing-free transformer architecture. Inspired by \cite{dndetr}, we developed IoU-guided Pseudo Label Refinement(IPLR) to evaluate box accuracy, filter and reweight pseudo-labels in conjunction with classification confidence score.

In IPLR, we introduce an IoU branch at the detection head, sharing the same structure as classification branch. After using Hungarian algorithm assigner\cite{enddetr}, we associate each ground truth with its predicted box and compute the IoU between them. This value is placed in the category position predicted, serving as the learning target. We employ varifocal loss \cite{varifocal} in this branch:
\vspace{-0.2cm}
\begin{equation}
    L_{vfl}(p,q)=\left\{
    \begin{aligned}
    & -q(q\log{(p)}+(1-q)\log{(1-p)}) \quad q>0 \\
    & -\alpha p^{\gamma}\log{(1-p)} \quad \quad \quad \quad \quad \quad \quad \quad q=0,
    \end{aligned}
    \right.
    \vspace{-0.2cm}
\end{equation}
where $p$ denotes the value predicted by the IoU branch, and $q$ signifies the actual IoU between the predicted box and the ground truth. It's worth noting that if the predicted box does not match the ground truth, the label predicted by the branch should be set to 0. $\alpha$ is weight coefficient and $\gamma$ is the focusing parameter. 

The predicted IoU values of the above branch serve as the localization certainty. We integrate classification confidence and localization certainty to get the combined confidence, as expresses by the following formula:
\vspace{-0.25cm}
\begin{equation}
    C = \sqrt{C_{class}\cdot C_{loc}}.
    \vspace{-0.2cm}
\end{equation}

Subsequently, we leverage the combined confidence to filter pseudo labels. Nevertheless, not all filtered labels possess equivalent credibility. To rectify it, we reweight them based on their combined confidence so as to assign more weight on the labels with higher credibility. The unsupervised target domain loss, post-reweighting, is expressed as follows:
\vspace{-0.25cm}
\begin{equation}
        L_{det}(X_{t},\hat{Y_{t}}) =\sum_{i}^{N}[(1+e^{C^{i}-1})(L_{cls}^{i}+L_{vfl}^{i})+e^{C^{i}-1}L_{reg}^{i}],
        \label{formular:formular6}
        \vspace{-0.25cm}
\end{equation}
where $L_{cls}$ and $L_{reg}$ represent the loss of classification and regression of the model Deformable DETR \cite{deformable}, respectively.

\begin{table}[t] 
\footnotesize
\centering
\setlength{\tabcolsep}{10pt} 
\setlength{\abovecaptionskip}{0cm}
\caption{Results of different methods on Sim10k to Cityscapes.}
\label{table2}
\begin{tabular}{l|l|l}
\toprule
Methods &Detector &Car mAP\\
\midrule
SWDA\cite{SWDA} &FRCNN &44.6  \\
ViSGA\cite{VisGA} & FRCNN &49.3  \\
PT\dag\cite{PT} & FRCNN &55.1  \\
MGA\dag\cite{MGA} &FRCNN &54.6  \\
\midrule
EPM\cite{EPM} &FCOS &47.3\\
KTNet\dag\cite{KTNet} &FCOS &50.7\\
\midrule
SFA\cite{SFA} &DefDETR &52.1\\
MTTrans\cite{MTTrans} &DefDETR &57.9\\
DA-DETR\cite{DA-DETR} &DefDETR &54.7\\
AQT\cite{AQT} &DefDETR &53.4\\
\midrule
ACCT(Ours) &DefDETR &\textbf{67.7}  \\
\bottomrule
\end{tabular}
\label{table:table2}
\vspace{-0.65cm}
\end{table}

\subsection{Category-aware Adaptive Threshold Generation}
\label{ssec:2.3}
Previous work \cite{category} has shown that using a single static threshold will reduce the accuracy and recall for minority classes. Inspired by \cite{semidetr}, we adopt the GMM module to generate adaptive category-specific thresholds. We consider the combined confidence of pseudo boxes predicted by the teacher network as the GMM model input, assuming the confidence distribution of each class is the sum of the gaussian distribution of the positive and negative modalities. Our GMM model is represented as follows:
\vspace{-0.25cm}
\begin{equation}
    P(C^{i}) = \sum_{j=1}^{K}\mathcal N(C^{i}|\mu_{j},\sigma_{j}^2),
    \vspace{-0.25cm}
\end{equation}
where $P(C^{i})$ represents the probability density of combined confidence of class i, K is set 2 due to two modalities and $\mathcal N(C^{i}|\mu_{j},\sigma_{j}^2)$ denotes the j-th gaussian distribution.

From the above modeling, we can estimate Gaussian distribution parameters and split the inputs into positive and negtive segment. We define the category thresholds as the minimum confidence score within the positive segment. During model training, predicted box scores may change their distribution across iterations. The GMM model can dynamically generate thresholds by learning new hybrid combinations.

\begin{table}[t]
\footnotesize
\centering
\setlength{\tabcolsep}{1.5pt}
\setlength{\abovecaptionskip}{0cm}
\caption{Results on adaptation from Cityscapes to BDD100K. The detector for each method is the same as Table 1 and Table 2.}
\label{table3}
\begin{tabular}{l|ccccccc|l} 
\toprule
Methods &person &rider &car &truck &bus &mcycle &bicycle &mAP\\
\midrule
SWDA\dag\cite{SWDA} &29.5 &29.9 &44.8 &20.2 &20.7 &15.2 &23.1 &26.2 \\
PT\dag\cite{PT} &40.5 &39.9 &52.7 &25.8 &33.8 &23.0 &28.8 &34.9 \\
\midrule
EPM\cite{EPM} &39.6 &26.8 &55.8 &18.8 &19.1 &14.5 &20.1 &27.8 \\
\midrule
SFA\cite{SFA} &40.2 &27.6 &57.5 &19.1 &23.4 &15.4 &19.2 &28.9\\
MTTrans\cite{MTTrans} &44.1 &30.1 &61.5 &25.1 &26.9 &17.7 &23.0 &32.6 \\
AQT\cite{AQT} &38.2 &33.0 &58.4 &17.3 &18.4 &16.9 &23.5 &29.4\\
\midrule
ACCT(Ours) &\textbf{51.8} &\textbf{41.4} &\textbf{61.8} &\textbf{26.0} &23.4 &\textbf{31.7} &\textbf{36.9} &\textbf{39.0}\\
\bottomrule
\end{tabular}
\label{table:table3}
\vspace{-0.65cm}
\end{table}

\subsection{Class-aware Contrastive Learning Module}
\label{ssec:subhead}
To counter data imbalance and improve object-level feature discrimination, we employ supervised contrast learning \cite{contrastive}. This encourages closer feature distributions among instances of the same category while pushing apart instances from different classes in the feature space. Our employed contrast learning loss function is defined as follows:
\vspace{-0.25cm}
\begin{equation}
    L_{contra} = \sum_{i \in I}-log \left\{\frac{w_{i}}{|P(i)|}\sum_{p\in P(i)}\frac{exp(z_{i}\cdot z_{p}/\tau)}{\sum_{\alpha \in A(i)}exp(z_{i}\cdot z_{\alpha}/\tau)}\right\},
    \vspace{-0.1cm}
\end{equation}
where $P(i)$ represents the objects from strong augmented images in student model of the same predicted class as object $i$, $A(i)$ includes all object of the different class from object $i$. $w_{i}$ is the weight of the object i.  $z_{i}$ denotes the feature of the object $i$ from weak augmented image in teacher model, extracted by a ROIAlign layer \cite{faster} with the multi-level feature map generated by cnn backbone in Deformable DETR, which can be expressed as follows:
\vspace{-0.2cm}
\begin{equation}
    z_{i} = ROIAlign(B_{i},F),
    \vspace{-0.2cm}
\end{equation}
where $B_{i}$ is the i-th filtered pseudo box, F means the feature maps.

However, previous studies \cite{contrastive} ignore confidence gaps among pseudo boxes. Lower-confidence boxes are more prone to label errors, which may narrow the feature distribution among different class instances. To address this, we reweight each object based on its combined confidence, similar to Equation \ref{formular:formular6}.

\begin{table}[t]
\small
\centering
\setlength{\tabcolsep}{2pt}
\setlength{\abovecaptionskip}{0cm}
\caption{Ablation studies on weather adaptation. AFL denotes the adversarial feature training and MT represents the Mean-Teacher learning. IPLR stands for the IoU-guided Pseudo Label Refinement and contra indicates the class-aware contrastive learning module.}
\begin{tabular}{l|ccccc|l} 
\toprule
Methods &MT &AFL &IPLR &gmm &contra &mAP\\
\midrule
baseline &\ding{53} &\ding{53} &\ding{53} &\ding{53} &\ding{53} &29.5 \\
AFL+MT&\checkmark &\checkmark &\ding{53} &\ding{53} &\ding{53} &43.2 \\
\midrule
\multirow{4}{*}{proposed}&\checkmark &\checkmark &\ding{53} &\ding{53} &\checkmark &44.6 \\
 &\checkmark &\checkmark &\ding{53} &\checkmark &\ding{53} &46.7 \\
 &\checkmark &\checkmark &\checkmark &\ding{53} &\ding{53} &47.9 \\
 &\checkmark &\checkmark &\checkmark &\checkmark &\ding{53} &48.9 \\
\midrule
ACCT(Ours) &\checkmark &\checkmark &\checkmark &\checkmark &\checkmark &\textbf{49.6} \\
\bottomrule
\end{tabular}
\label{table:table4}
\vspace{-0.65cm}
\end{table}

Data imbalance poses challenges for certain classes. In section \ref{ssec:2.3}, we generate class-specific thresholds, where lower threshold imply more difficult learning. To handle this, we allocate higher weights to challenging category samples. In summary, an object's weight is determined by both combined confidence and class threshold as follows:
\vspace{-0.2cm}
\begin{equation}
    w_{i} = \frac{(1+e^{C^{i}-1})(1-\tau_{j}^{\gamma})}{\sum_{i}^{N}(1+e^{C^{i}-1})(1-\tau_{j}^{\gamma})},
    \vspace{-0.2cm}
\end{equation}
where $N$ denotes the number of the objects, and $\tau_{j}$ present the threshold of the class predicted by object i and $\gamma$ is set 0.5.
\subsection{Overall Loss}
\label{ssec:loss}
In the burn-up stage, the loss contains the adversarial loss and detection loss from source domain. The loss is as follows:
\vspace{-0.25cm}
\begin{equation}
    L_{burn} = L_{det}(X_{s},Y_{s}) -\lambda_{adv}L_{adv}.
    \vspace{-0.25cm}
\end{equation}

In the mutual learning stage, the loss contains the student model loss, adversarial loss and contrastive loss. The formula is as follows:
\vspace{-0.25cm}
\begin{equation}
    L_{multual} = L_{s}+\lambda_{contra}L_{contra}-\lambda_{adv}L_{adv}.
    \vspace{-0.25cm}
\end{equation}

\section{EXPERIMENT}
\label{sec:pagestyle}
\subsection{Experiment Setup}
\label{ssec:subhead}
\textbf{Dataset.} We evaluate our method on four domain adaptation scenarios: weather adaptation (Cityscapes to Foggy Cityscapes), synthetic to real world adaptation (Sim10k to Cityscapes), and scene adaptation (Cityscapes to BDD100K).\\
\textbf{Implementation Details.} Our method builts upon Deformable DETR following the \cite{MTTrans}'s experimental setting. In the burn-in step, we use Adam optimizer for 50 epochs of training with batch size of 4 for all tasks. The initial learning rate is $2\times10^{-4}$, decreasing by 0.1 after 40 epochs. $\lambda_{adv}$ is 1 for weather adaptation and 0.1 otherwise. In the mutual learning phase, we train for 40 epochs with batch size 2, starting at $2\times10^{-5}$ learning rate, reduced by 0.1 after 20 epochs. $\lambda_{contra}$ is set to 0.05 for all experiments. We report the mean Average Precision(mAP) with an IoU threshold of 0.5 for the teacher network during testing.
\begin{figure}[t]
	\centering  
	\vspace{-0.35cm}
 	\setlength{\abovecaptionskip}{-0.2pt} 
	\subfigbottomskip=0.5pt 
	\subfigcapskip=-3pt 
	\subfigure[Ground Truth]{
		\includegraphics[width=0.48\linewidth]{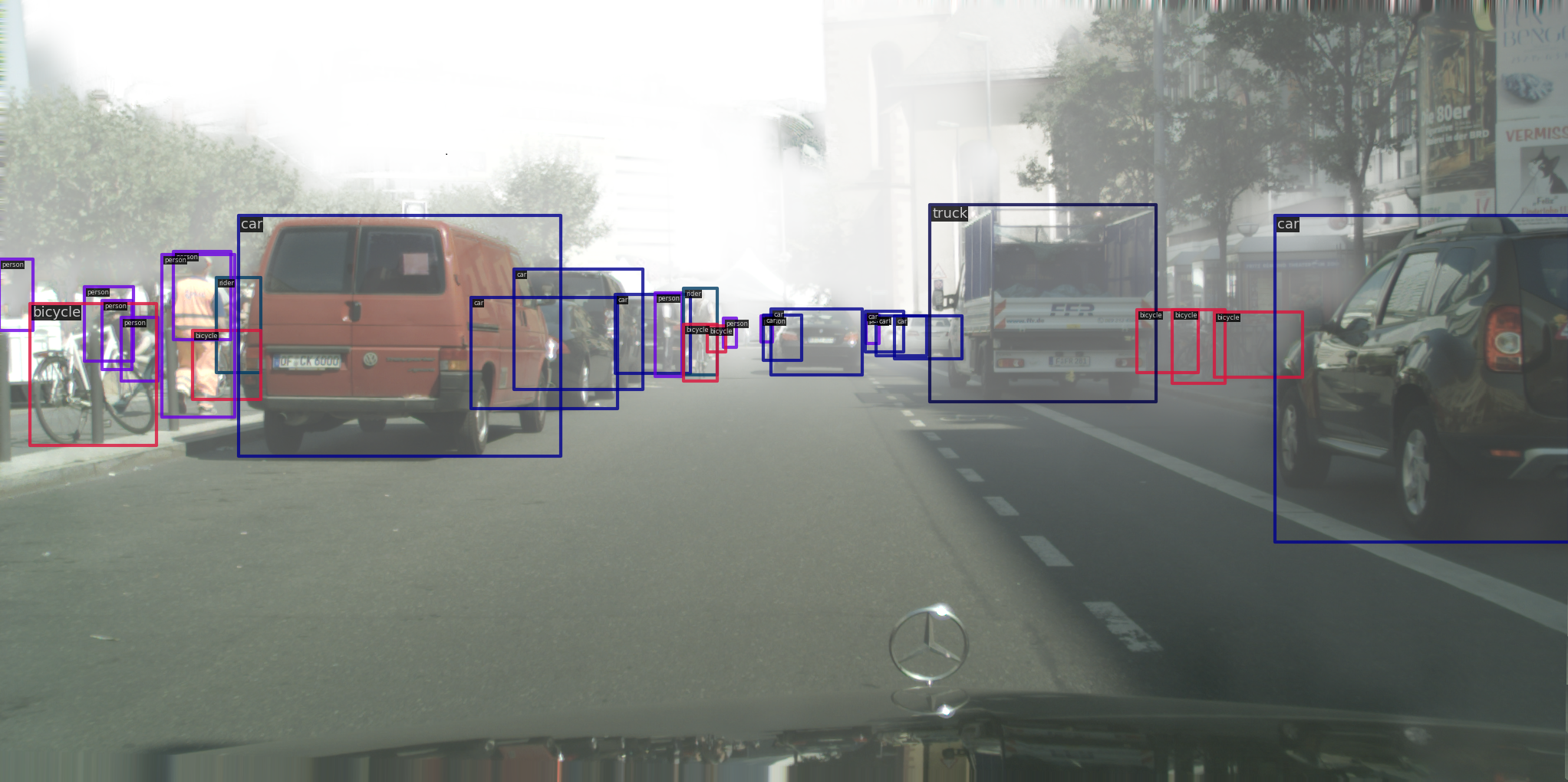}}
	\subfigure[SFA]{
		\includegraphics[width=0.48\linewidth]{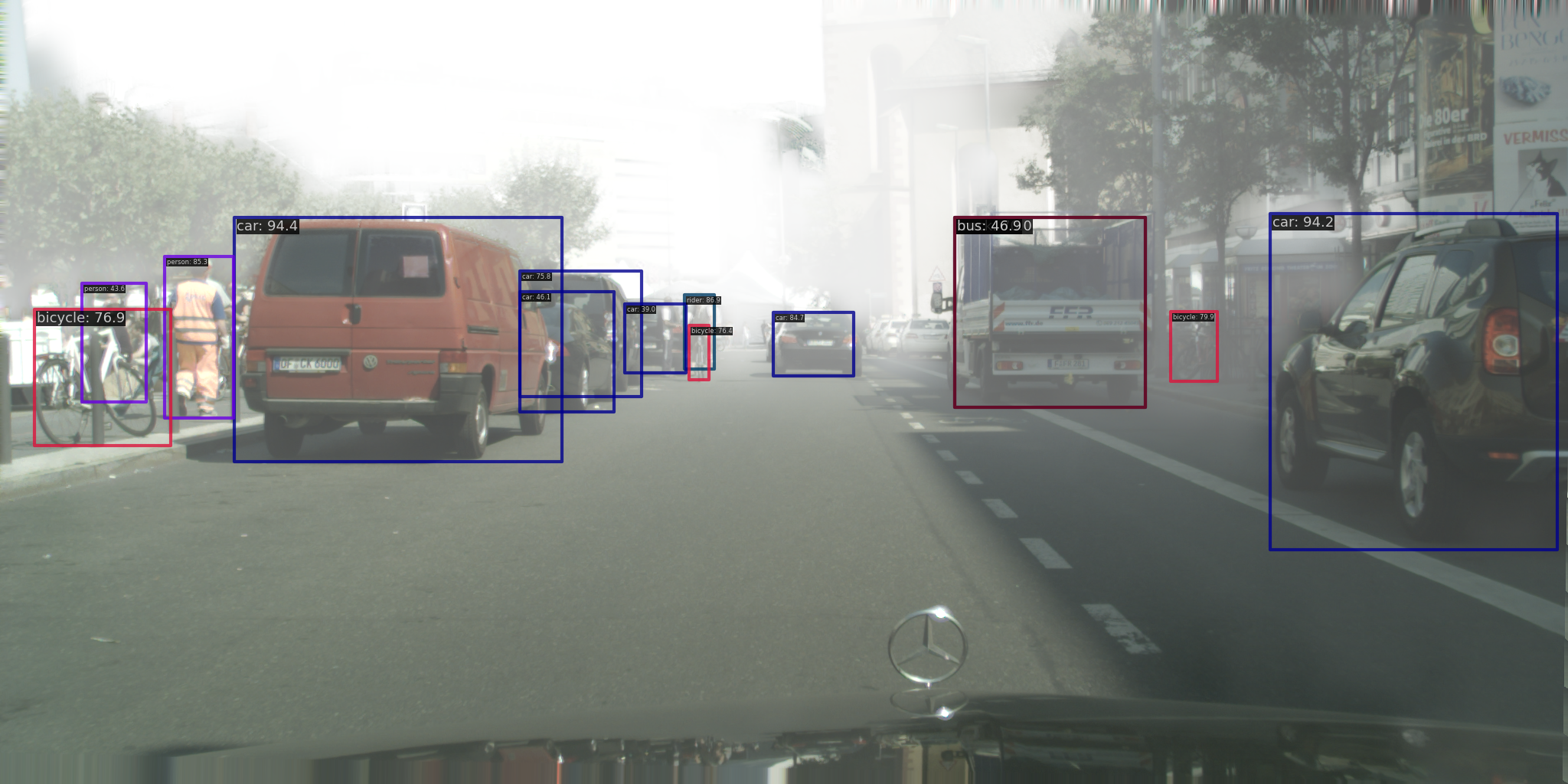}}
	  \\
	\subfigure[MTTrans]{
		\includegraphics[width=0.48\linewidth]{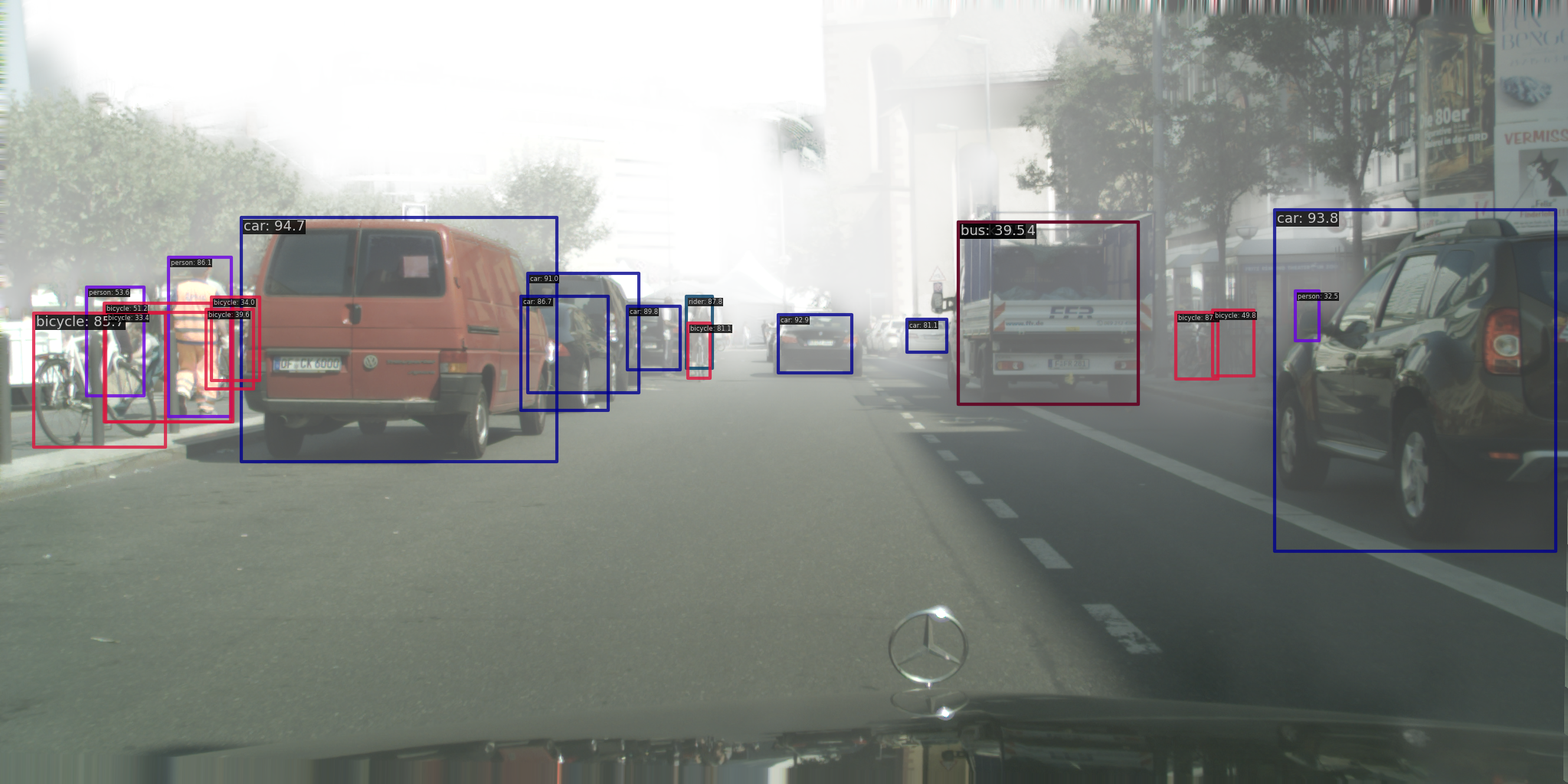}}
	\subfigure[Ours]{
		\includegraphics[width=0.48\linewidth]{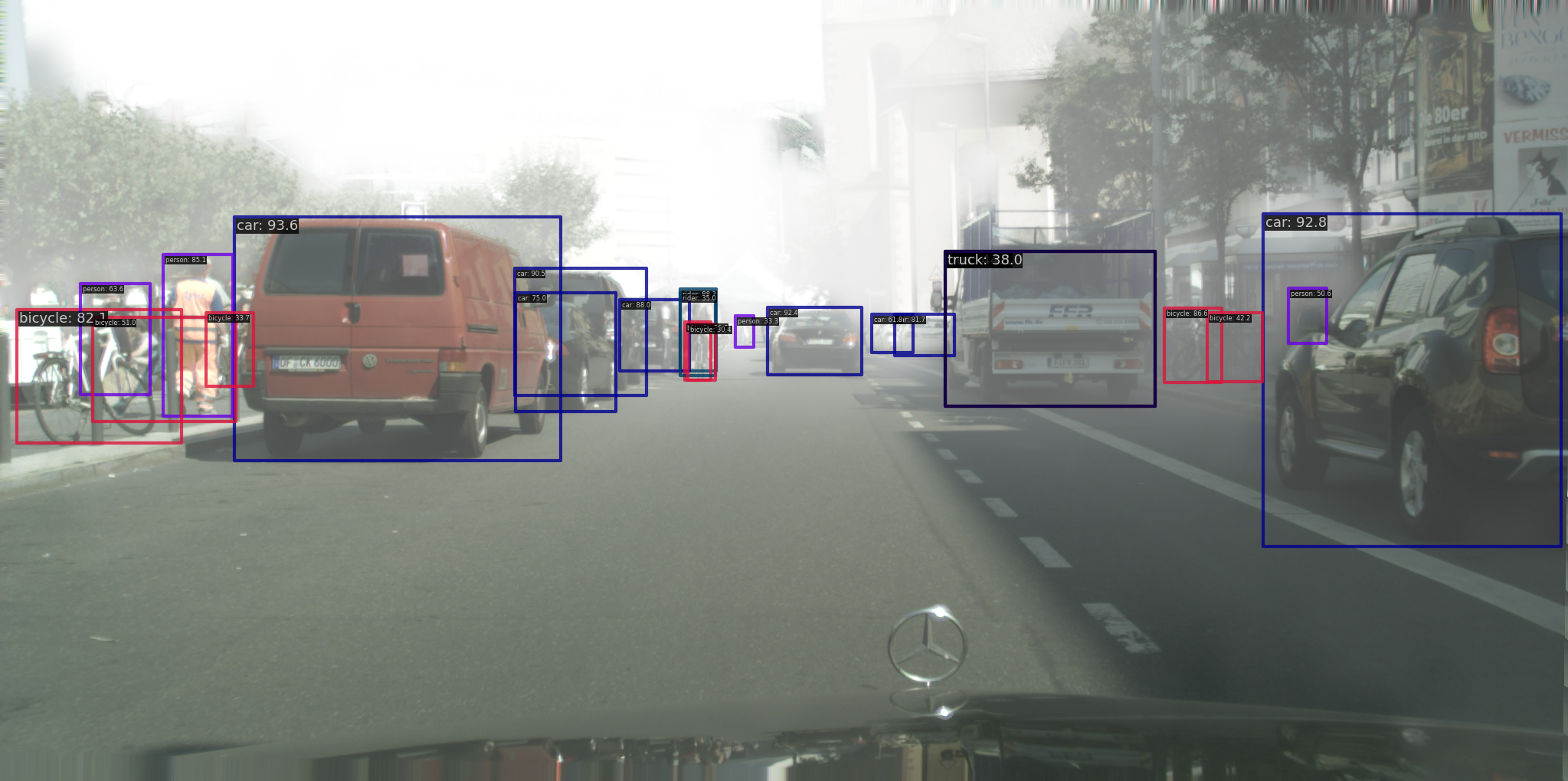}}
	\caption{Visualizations on the weather adaptation of different methods based transformer}
	\label{fig:vis}
    \vspace{-0.5cm}
\end{figure}
\begin{figure}[t]
	\centering  
	\vspace{-0.1cm}
 	\setlength{\abovecaptionskip}{-0.2pt} 
	\subfigbottomskip=0.5pt 
	\subfigcapskip=-3pt 
	\subfigure[Baseline]{
		\includegraphics[width=0.46\linewidth]{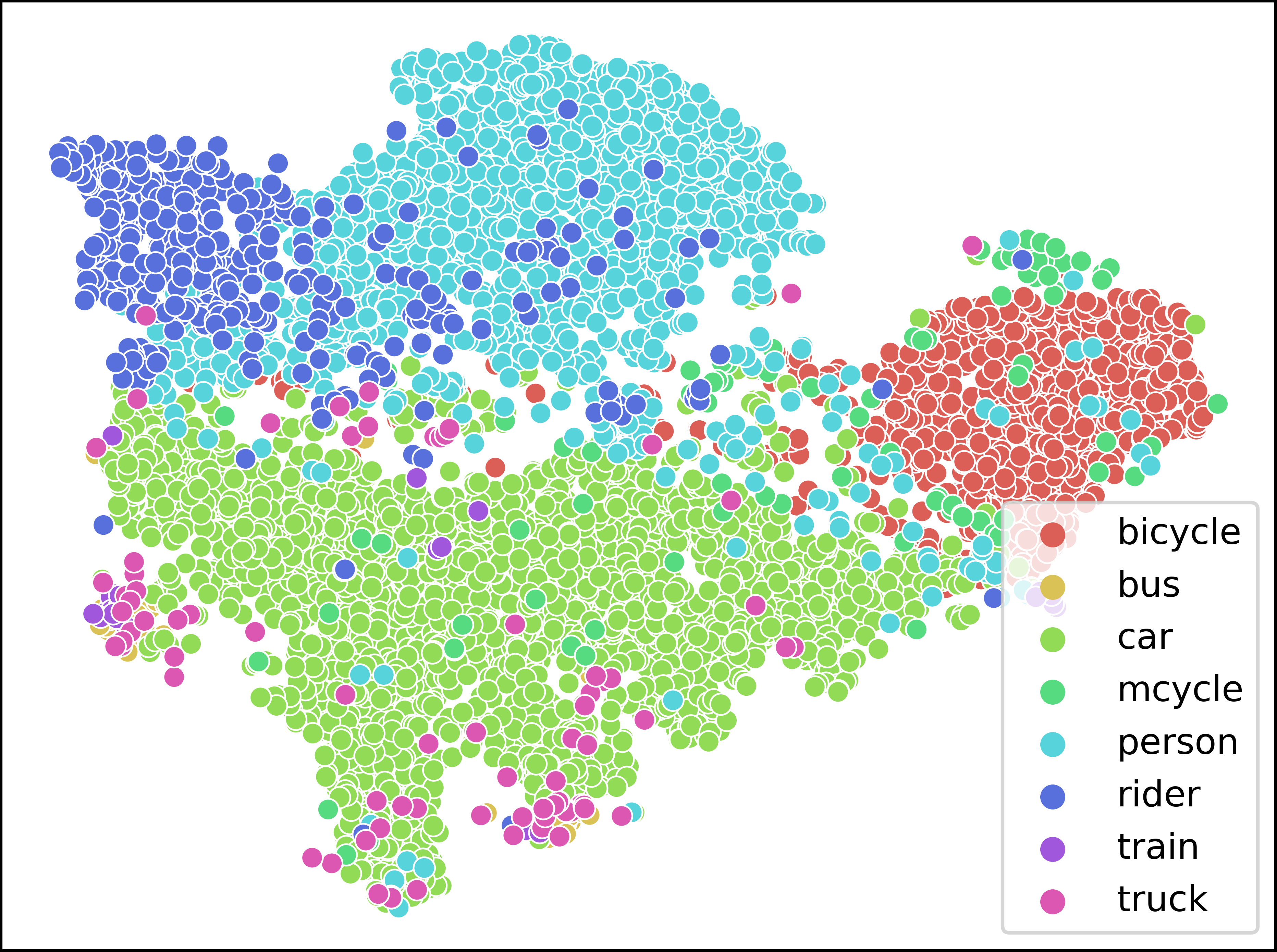}}
	\subfigure[Class-aware contrastive module]{
		\includegraphics[width=0.46\linewidth]{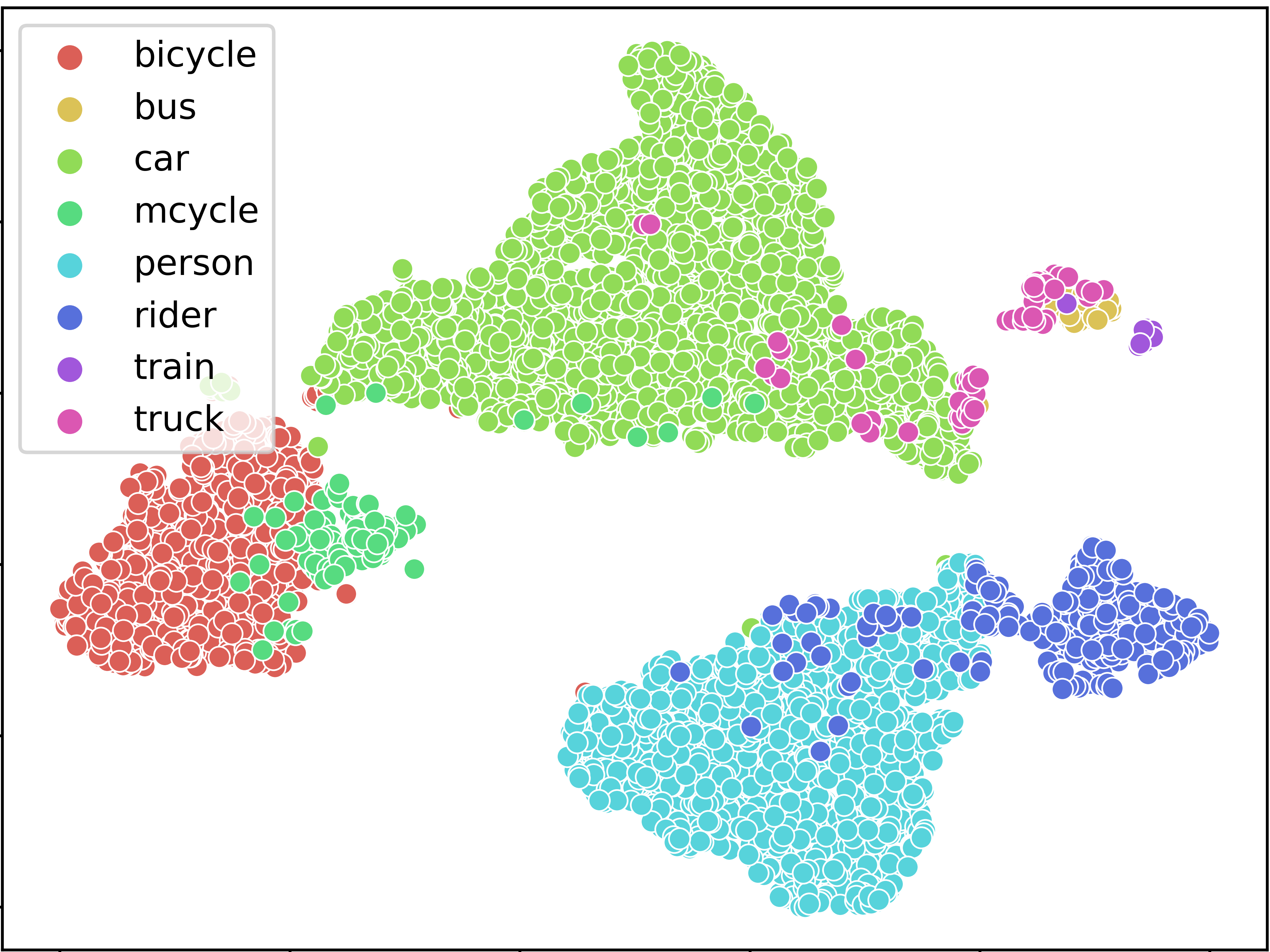}}
	\caption{Visualization of foggy cityscapes val dataset in t-sne }
	\label{fig:t-sne}
    \vspace{-0.7cm}
\end{figure}
\subsection{Comparison with the State-of-the-Art}
\label{ssec:subhead}
\textbf{Weather Adaptation.} As shown in Table \ref{table:table1}, our method outperforms other transformer based methods by a significant margin, especially for some minority classes such as \textbf{truck} and \textbf{motorcycle}, proving that our approach indeed helps mitigate the issue of class imbalance.\\
\textbf{Synthetic to Real Adaptation.} In Table \ref{table:table2}, we can observe that our method reaches an mAP of $67.7\%$ with a gain of \textbf{$+9.8\%$} over the previous state-of-the-art method.\\
\textbf{Scene Adaptation.} As shown in Table \ref{table:table3}, our method achieves a state-of-art performance in almost all classes, which reflects that our method can alleviate the negative effects of the class imbalance.
\subsection{In-depth Analysis}
\label{ssec:subhead}
\textbf{Ablation Study.} We conduct ablation studies on the Weather Adaption datasets in Table \ref{table:table4}to assess our proposed components. Our strong baseline, AFL+MT, integrates Deformable DETR with adversarial feature learning and the mean-teacher framework with threshold for 0.5. In our experiments, each component has an individual performance improvement over the baseline. In addition, integrating all components, our model achieved the best performance, which demonstrates that our proposed components are complementary to each other.\\
\textbf{Qualitative Visualization.} In Fig \ref{fig:vis}, we illustrate detection examples for different methods in the weather adaptation task. Compared to other methods, Our method excels in recognizing distant and small objects and demonstrates superior performance in correctly identifying minority classes like trucks. In Figure \ref{fig:t-sne}, we use t-sne to visualize features extracted by ROIAlign for all objects in the foggy cityscapes val dataset. We establish a baseline by removing the contrastive module from our model. Intuitively, the contrastive module enhances feature separation and facilitates the network in learning more discriminative features especially for minority class.
\section{CONCLUSION}
\label{sec:majhead}
In this paper, we explore the constraints of inaccurate pseudo boxes and biased class monitoring in existing domain adaptation object detection methods. To improve localization ability, we introduce an IPLR module, which combines localization and classification scores to filter pseudo labels. To tackle the class-bias problem, we propose a GMM model to dynamically generate class-specific thresholds and a reweighted contrast learning module to enhance minority class performance. Experimental results on three different adaptation tasks validate the effectiveness and superiority of our method.

\vfill\pagebreak



\bibliographystyle{IEEEbib}
\bibliography{strings,refs}

\end{document}